\documentclass{article}

\usepackage[preprint]{corl_2020} % Uncomment for pre-prints (e.g., arxiv); This is like ``final'', but will remove the CORL footnote.

\usepackage{comment}
\usepackage{graphicx}
\usepackage{amsmath,amssymb} % define this before the line numbering.
\usepackage{color}
\usepackage{wrapfig}
\usepackage[table]{xcolor}
\usepackage{booktabs}
\newcommand\mypara[1]{\vspace{1mm}\noindent\textbf{#1}}
\usepackage{array}
\newcolumntype{M}[1]{>{\centering\arraybackslash}m{#1}}
\usepackage{hyperref}
\usepackage{url}
\usepackage{booktabs}
\usepackage{multirow}

\newcommand{\ie}{{\em i.e.}}

\newcommand{\eg}{{\em e.g.}}

\newcommand{\Fig}[1]{Fig. \ref{fig:#1}}

\newcommand{\SSect}[1]{Sect. \ref{ssec:#1}}

\newcommand{\Du}{\mathcal{D}^u}
\newcommand{\Dl}{\mathcal{D}^l}
\newcommand{\superscriptOne}[1]{#1\ensuremath{^1}\kern-\scriptspace}
\newcommand{\superscriptTwo}[1]{#1\ensuremath{^2}\kern-\scriptspace}

\title{Action-based Representation Learning for Autonomous Driving}

\author{
  \superscriptOne{Yi Xiao}, \superscriptTwo{Felipe Codevilla}, \superscriptTwo{Christopher Pal}, \superscriptOne{Antonio M. L\'opez}\\
  \\
  \superscriptOne{}Computer Vision Center (CVC) and Computer Science Dpt. \\ 
                  at the Universitat Autonoma de Barcelona (UAB), Barcelona, Spain\\
  \superscriptTwo{}Montreal Institute for Learning Algorithms (MILA), Montreal, Canada\\
  }

\begin{document}
\maketitle

\begin{abstract}
Human drivers produce a vast amount of data which could, in principle, be used to improve autonomous driving systems. Unfortunately, seemingly straightforward approaches for creating end-to-end driving models that map sensor data directly into driving actions are problematic in terms of interpretability, and typically have significant difficulty dealing with spurious correlations. Alternatively, we propose to use this kind of action-based driving data for learning representations. Our experiments show that an affordance-based driving model pre-trained with this approach can leverage a relatively small amount of weakly annotated imagery and outperform pure end-to-end driving models, while being more interpretable.
Further, we demonstrate how this strategy outperforms previous methods based on learning  inverse dynamics models as well as other methods based on heavy human supervision (ImageNet).
\end{abstract}


%===============================================================================
\linespread{1}

\section{Introduction} 
%Autonomous vehicles (AVs) are the key enabling technology for future mobility.
The development of autonomous vehicles (AVs) is a significant multidisciplinary challenge. Currently, the main paradigm being pursued for developing AVs follows a traditional divide-\&-conquer engineering strategy. In particular, modular pipelines are proposed with key modules for perception, route planning and maneuver control, among others \cite{Yurtsever:2019}. In turn, these modules may be composed to deal with different tasks, {\eg}, perception encompasses object detection and tracking, semantic class/instance segmentation, etc. \cite{Janai:2019}. These tasks rely on models trained from data using modern deep learning techniques \cite{Grigorescu:2019}. Following such a data-driven approach is not a problem in itself since it is possible to collect petabytes of on-board data (raw sensor data, vehicle state variables, etc.) continuously, not only from fleets of AVs under development, but also from sensorized human-driven vehicles under naturalistic driving. However, in practice, the best performing models arise from supervised deep learning, and this means that the raw data must be augmented with ground truth, which is collected through time consuming and costly human annotations ({\eg}, bounding boxes, object silhouettes, etc). 

The data annotation bottleneck associated with these approaches has caused the idea of end-to-end driving \cite{Pomerleau:1989,LeCun:2005} to receive renewed interest \cite{Bojarski:2016,Xu:2017,Hecker:2018,Amini:2018,Codevilla:2018,Codevilla:2019}.  In this paradigm a deep model is trained to directly control an AV from input raw sensor data (mainly images), {\ie}, without a clear separation between perception and maneuver planning, and without explicit intermediate perceptual tasks to be solved. In this pure data-centered approach, the supervision required to train deep end-to-end driving models does not come from 
human annotation; instead, the vehicle's state variables, which can be automatically collected from fleets of human-driven vehicles, are used as supervision ({\eg} speed, steering angle, acceleration, braking). 
% Before: instead, the vehicle's state variables are used as self-supervision ({\eg} speed, steering angle, acceleration, braking) since these can be automatically collected from fleets of human-driven vehicles.
These models are mainly trained by behaviour cloning (BC) of human driving experiences. However, despite the undeniable good performance shown by end-to-end driving models, their reliability is controversial, due in particular to the difficulty of interpreting the relationship between inferred driving actions and image content \cite{Levine:2018}, as well as training instabilities \cite{Codevilla:2019}. 
% Antonio's comment: from the intro I have skipped these references (since I did not have time to check them, it is not clear for me what "drawback" of e-2-e driving models they focus on) => shalev2016sample, kim2018textual, zhou2019does, sax2018mid.

% Felipe's comment: I think the first paragraphs are overly explanatory we arrive in our objective only on paragraph 4.
A different paradigm, conceptually midway between pure modular and end-to-end driving, is the so-called direct perception approach \cite{Chen:2015,Sauer:2018}, which focuses on learning deep models to predict driving affordances, from which an additional controller can maneuver the AV. In general, such affordances can be understood as a relatively small set of interpretable variables describing events that are relevant for an agent acting in an environment \cite{Gibson:2014}. Driving affordances bring interpretability while only requiring weak supervision, in particular, human annotations just at the image level ({\ie}, not pixel-wise).      

\begin{figure}[t!]
	\centering
    \begin{tabular}{c@{\hspace{5mm}}c}
	  \includegraphics[height=2.05cm]{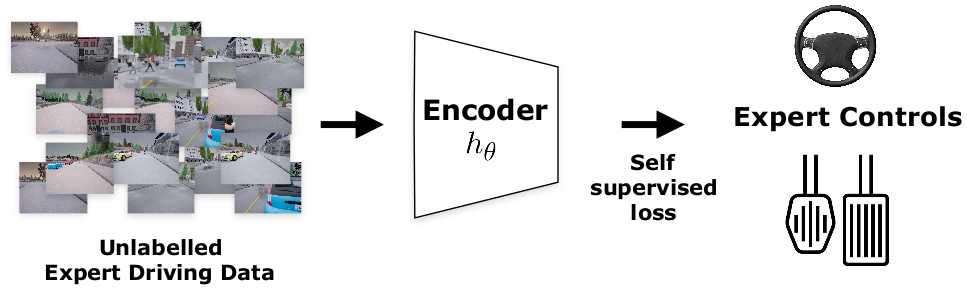} & 
	  \includegraphics[height=1.55cm]{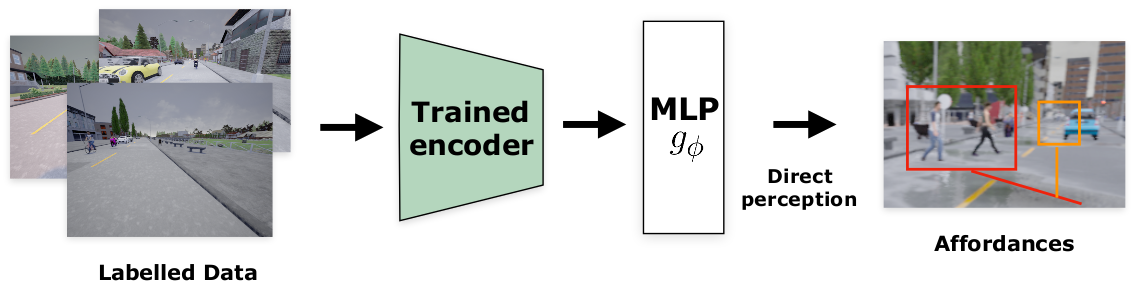}  \\
      {\tiny (a) Action-based supervised training stage 
      % Before: Self-supervised training stage 
      (end-to-end driving)} & 
      {\tiny (b) Weakly supervised training stage (direct perception)}
    \end{tabular}
    \caption{Approach overview: (a) an encoder is trained following an end-to-end driving setting ({\eg} using BC or inverse model); (b) this pre-trained encoder together with a multi-layer perceptron (MLP) are used for predicting affordances. The affordances are used as input to a simple PID controler to drive the vehicle.}
	\label{fig:general_arch}
  \vspace{-3mm}
\end{figure}

In this paper, we show that action-based methods, that focus on predicting the control
actions, such as end-to-end driving trained with BC, can be an effective pre-training strategy for learning a direct perception model (\Fig{general_arch}). This strategy enables a significant reduction on the number of annotated images required to train such a model. Overall, this means that we can leverage the data 
% Before: self-supervised data
collected by fleets of human-driven vehicles for training interpretable driving models, thus, keeping a major advantage of modular pipelines while reducing data supervision ({\ie}, human annotation). Further, our approach, improves over other recent %\ToBeReviewed{supervised} 
% Before: self-supervised
pre-training proposals such as contrastive methods \cite{Anand:2019} and even over ImageNet (supervised)
%\ToBeReviewed{deleted ``(supervised)"} 
pre-training. We also show that learning from expert data in our approach leads to better representations compared to  training inverse dynamics models using the approach in \cite{Agrawal:2016}. This shows that expert driving data ({\ie} coming from human drivers) is an important source for representation
learning. As is common practice nowadays, we run our experiments in the 
CARLA simulator \cite{Dosovitskiy:2017}. 
To the best of our knowledge, this work is the first to show that expert demonstrations can act as an effective action-based representation learning technique. This constitutes the primary contribution of this paper.

\section{Related Work}
\label{sec:rw}
Since human-based data annotation is a general problem for all kinds of new data-intensive applications, not only for autonomous driving, learning representations (deep models) with the support of weak supervision and self-supervision are open challenges that attract great interest. 

In the autonomous driving context, the use of driving affordances \cite{Chen:2015,Sauer:2018} allows for weak supervision since only annotations at the image level are required. Based on these interpretable affordances a controller is tuned to drive. Since \cite{Sauer:2018} focuses on urban driving using CARLA simulator, inspired from this work, we have defined four affordances to consider the explicit detection of hazards involving pedestrians and vehicles, respecting traffic lights and considering the heading of the vehicle within the current lane. Defining the best set of affordances to drive is not the focus of this paper, but we have chosen a reasonable set. 

%For instance, using CARLA simulator to develop a deep driver, in \cite{Sauer:2018} there are considered three discrete and three continuous variables (affordances); in short, \emph{hazard stop} (t/f), \emph{red traffic light} (t/f), \emph{speed sign} (four classes), \emph{distance to preceding vehicle} (m), \emph{relative vehicle heading} (rad), and \emph{on-lane vehicle lateral position} (m). Based on these interpretable affordances a controller is tuned to drive. After insights from preliminary experiments, we have focused this work on using four affordances (\Fig{affordances}), namely, \emph{pedestrian hazard} (t/f), \emph{vehicle hazard} (t/f), \emph{red traffic light} (t/f), and \emph{relative vehicle heading} (rad).

In order to solve visual tasks, we can find self-supervision
based on auxiliary and relatively simple (pretext) tasks such as learning colorization \cite{Larsson:2017}, rotations \cite{Gidaris:2018,Feng:2019,Xu:2019}, shuffling cues \cite{Misra:2016}, or solving a jigsaw puzzle of image parts \cite{Noroozi:2016}; it has been shown that self-supervision can match traditional ImageNet (supervised) pre-training provided one works with large enough CNNs \cite{Kolesnikov:2019}, although it has been argued that these proxy tasks are not sufficiently hard so as to fully exploit large unsupervised datasets \cite{Goyal:2019}. Another branch of self-supervised learning is based on contrastive methods \cite{Hadsell:2006,Oord:2018,Chen:2020}, which learn representations by comparing data pairs. While, in these methods supervision
% self-supervision
is based on different ways of transforming or comparing the input data itself, in this paper, 
supervision 
% Before: self-supervision 
comes in the form of expert driver actions. In fact, we include in our study a recent contrastive method, ST-DIM, designed in the context of playing Atari games \cite{Anand:2019}, adapted to actions required for driving. We will see how action-based representation learning outperforms ST-DIM as a %\ToBeReviewed{deleted ``self-supervised"} 
% Before: self-supervised
representation learning strategy to infer affordances. 

In fact, in a perceive-\&-act context, dynamics learning \cite{Chua:2018,Hafner:2019} and inverse dynamics \cite{Agrawal:2016,Shelhamer:2017,Pathak:2017} can be used as action-based supervision 
% before: self-supervision
strategy. Broadly speaking, being able to predict the next states of an agent, or the action between state transitions, yields useful representations. This action-centric approach to supervision 
% Before: self-supervision
is in line with our work. Thus, our study includes experiments with different inverse and forward dynamics supervision
% Before: self-supervision
strategies. We show the importance
of those strategies  in the autonomous driving context. However, different than previous work, we empirically demonstrate that %\ToBeReviewed{deleted ``self-supervised"} 
expert actions yields a better representation learning than random actions used in \cite{Agrawal:2016}.

Finally, it is also worth mentioning teacher-student strategies \cite{Chen:2019,Zhao:2019} which allow one to train an end-to-end driving student model from a teacher model. In this case, even the student is end-to-end, the data annotation bottleneck arises during the supervised training of the teacher, which requires bounding boxes and/or semantic segmentation. Since the student is still an end-to-end driving model, the issue of interpretability once this model is deployed in the AV still would remain open. %In this paper, 
%learning by cheating \cite{Chen:2019} and conditional behaviour cloning \cite{Codevilla:2019}
%the approach in \cite{Chen:2019} will be used as the driving reference, but our focus is on the prediction of affordances themselves by behaviour cloning; thus, to compare our affordances with such baselines when driving, we will just use a relatively simple controller to convert affordances into vehicle actions. 

%%%%%%%%%%%%%%%%%%%%%%%%%%%%%%%%%%%%%%%%%%%%%%%%%%%%%%%%%%%%%%%%%%%%%%%%%%%%%%%%%%%%%%%%%%%%%%%%%%%%%%%%%%%%%%%%%%%%%%%%%%%%%%%%%%%%%%%%%%%%%%%%%%%%%%%%%%%%%%%%%%%%%%%%%%%%%%%%%%%%%%%%%%%%%%%%%%%%%%%%%%%%%%%%%%%%%%%%%%%%%%%%%%%%%%%%%%%%%%%%%%%%%%%%%%%%

\begin{figure}[t]
 \centering
  \includegraphics[width=0.95\linewidth]{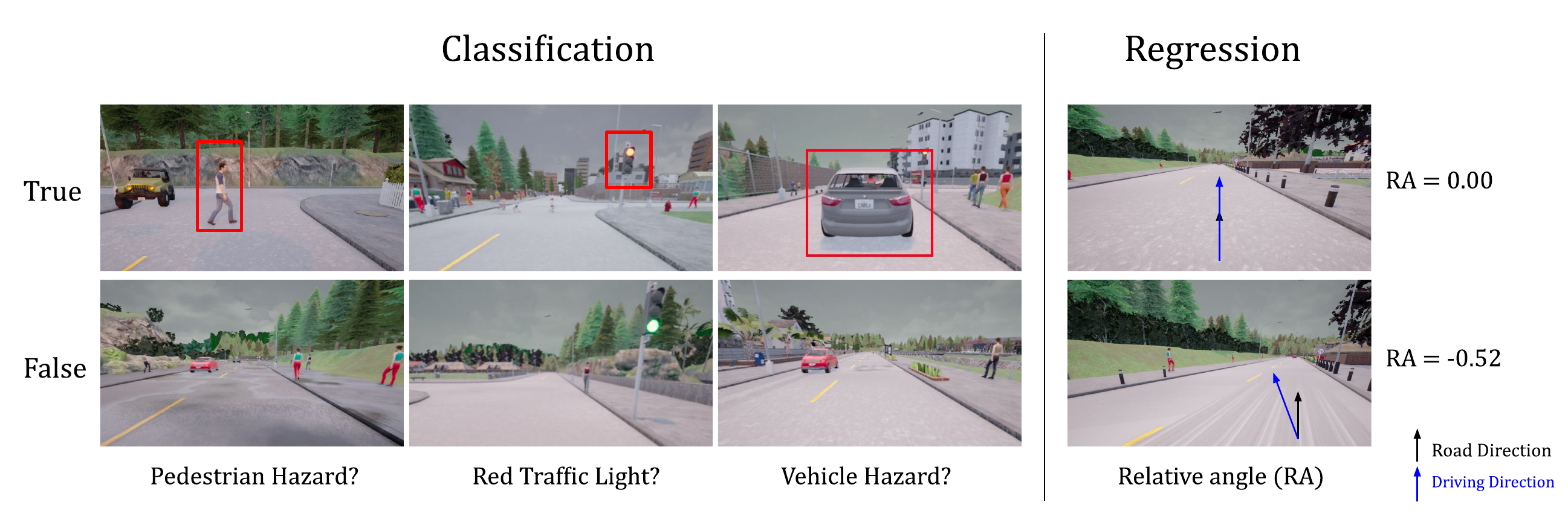}
  \caption{Our affordances illustrated on images from CARLA. Classification ones are binary variables (t/f), and regression runs on $[-\pi,\pi]$rad. See \SSect{a} for details.}
  \label{fig:affordances}
\end{figure}

\section{Action-based Representation Learning}
\label{sec:arl}

\subsection{Overall Approach}
\label{ssec:oa}
As can be seen in \Fig{general_arch}, we study our
action-based representation learning strategy  by
learning affordances in two stages. The first stage relies on 
non-manually annotated data 
% Before: self-supervised data
to learn a representation (encoder). We will consider different methods to learn this representation (\SSect{sss}), all of them based on predicting driving actions from on-board data. The second stage uses this pre-trained representation together with a multi-layer perceptron (MLP) to learn the considered affordances (\SSect{a}).
%; where we will study the case of freezing the pre-trained representation and the case of fine-tuning it. 

Therefore, for the first stage of our approach, we assume that we have access to a sequence of $N_u$ data samples $\Du=\{d_t\}_{t=1}^{N_u}$, which have been acquired on-board a human-driven sensorized vehicle but without
%they have no 
human annotations. 
%\ToBeReviewed{(deleted ``and will be used as self-supervised data")}. 
Thus, we have $d_t=\{o_t,a_t\}$, where $o_t$ and $a_t$ are respectively, at a certain time $t$, the observation acquired by the vehicle's sensors and the driving action taken by the expert driver. We must understand that $a_t$ is the expert reaction to the environment when $o_t$ was acquired. In general, we will have different ${\Du}$ sequences acquired at different driving runs, however, without losing generality and for the sake of keeping a simpler notation, we can assume all of them appended in one single sequence. In this paper, we assume that each observation $o_t$ contains an \emph{image} capture of the driving environment, the \emph{vehicle speed} ($v_t$) at the moment the image is acquired, and a high level \emph{navigation command} such as {\tt continue} in the same lane, or in the next intersection go {\tt straight/left/right}, {\ie} as introduced in the so-called conditional imitation learning \cite{Codevilla:2018} in the form of one-hot vector $c_t$. The corresponding action $a_t$ is defined in terms of the \emph{steering angle}, \emph{acceleration}, and \emph{break} values that must be applied to maneuver the vehicle.

For the second stage of our approach, we assume a relatively small dataset of on-board images with image-level affordance annotations (weak supervision), {\ie}, $\Dl=\{x_j\}_{j=1}^{N_l}$, with $x_j=\{o_j,y_j\}$, being $o_j$ the observation and $y_j$ the corresponding affordance annotation. In particular, $y_j$ contains variables indicating situations such as a \emph{pedestrian hazard}, a \emph{vehicle hazard}, a \emph{red traffic light}, and a \emph{relative heading angle} (\Fig{affordances}). In this setting, we can assume that the images used in $\Dl$ come from sub-sequences of $\Du$, but selected so that $N_u~\!\!\!\ggg\!\!\!~N_l$.% Note that ${\Dl}$ does not include action-related information. 

Accordingly, the two stages can be summarized as follows: (1) use $\Du$ to train a deep encoder $h_\theta$; (2) use $h_\theta$ and $\Dl$ to train a projection network, $g_\phi$, for predicting affordances. For actual driving,  we develop a controller $C:g_\phi(h_\theta(o_t))\rightarrow\hat{a}_t$; {\ie}, given the affordances $g_\phi(h_\theta(o_t))$ predicted from an observation $o_t$, $C$ estimates the action $\hat{a}_t$ to maneuver the vehicle. In order to show driving results, we will use a simple PID controller.

\begin{figure}[t]
	\centering
    \begin{tabular}{M{0.5cm}M{1.5cm}M{5.5cm}M{3.7cm}}
    \toprule
          & Method                & Architecture                                & Loss \\
	  \midrule
	  (a) & Behavior Cloning (BC) & \includegraphics[height=1.3cm]{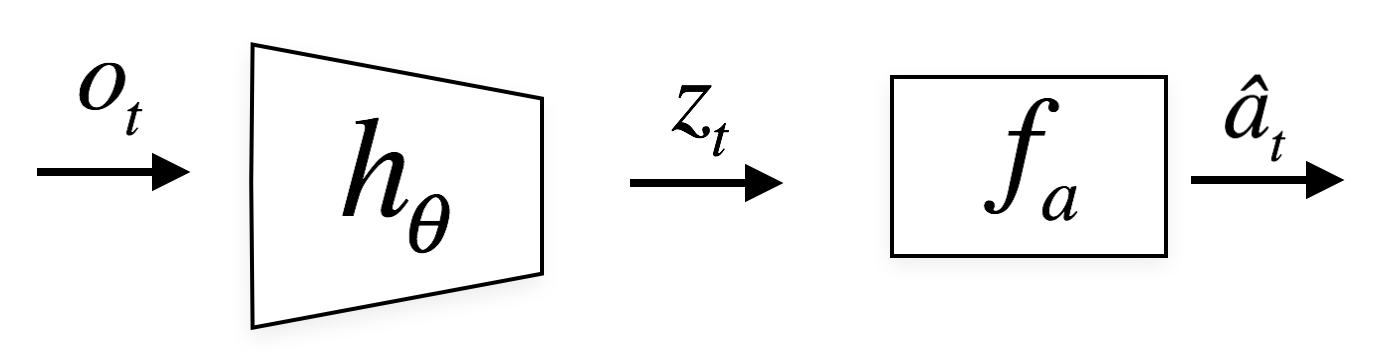}     &  $s_{bc} = ||f_a(z_t) - a_t ||$ \\
	  \midrule
      (b) & Inverse Model         & \includegraphics[height=1.7cm]{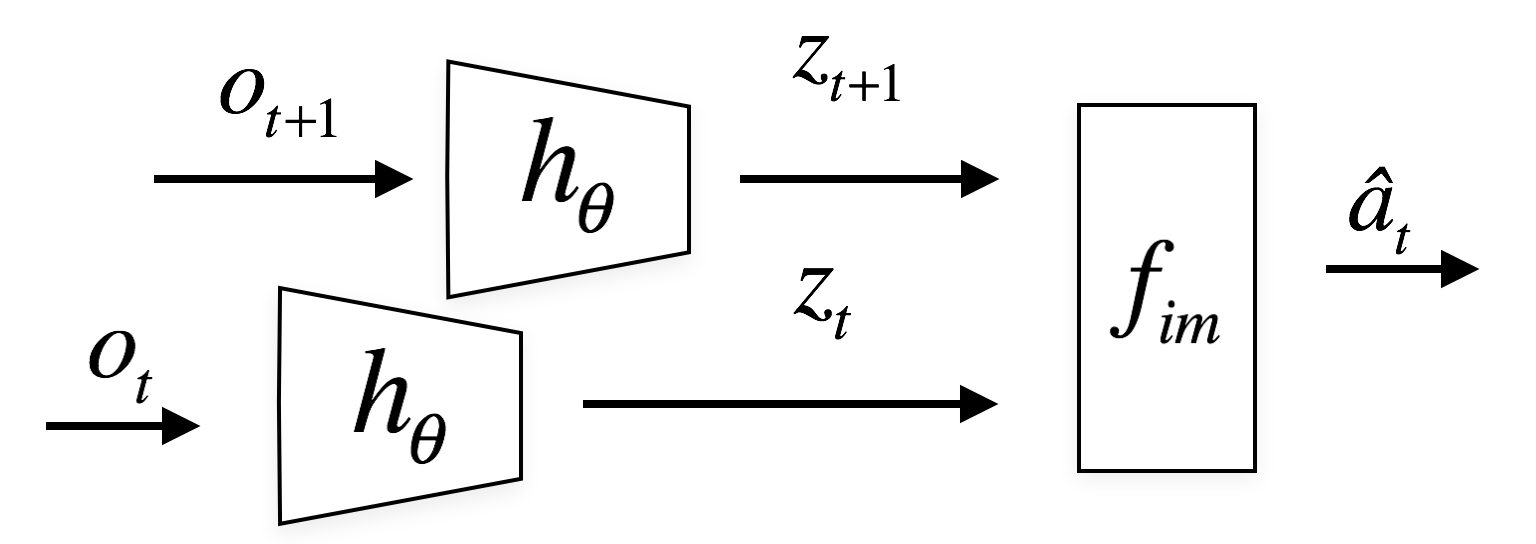}     & $s_{im} = ||f_{im}(z_t,z_{t+1}) - a_t ||$ \\
      \midrule
      (c) & Forward Model         & \includegraphics[height=1.7cm]{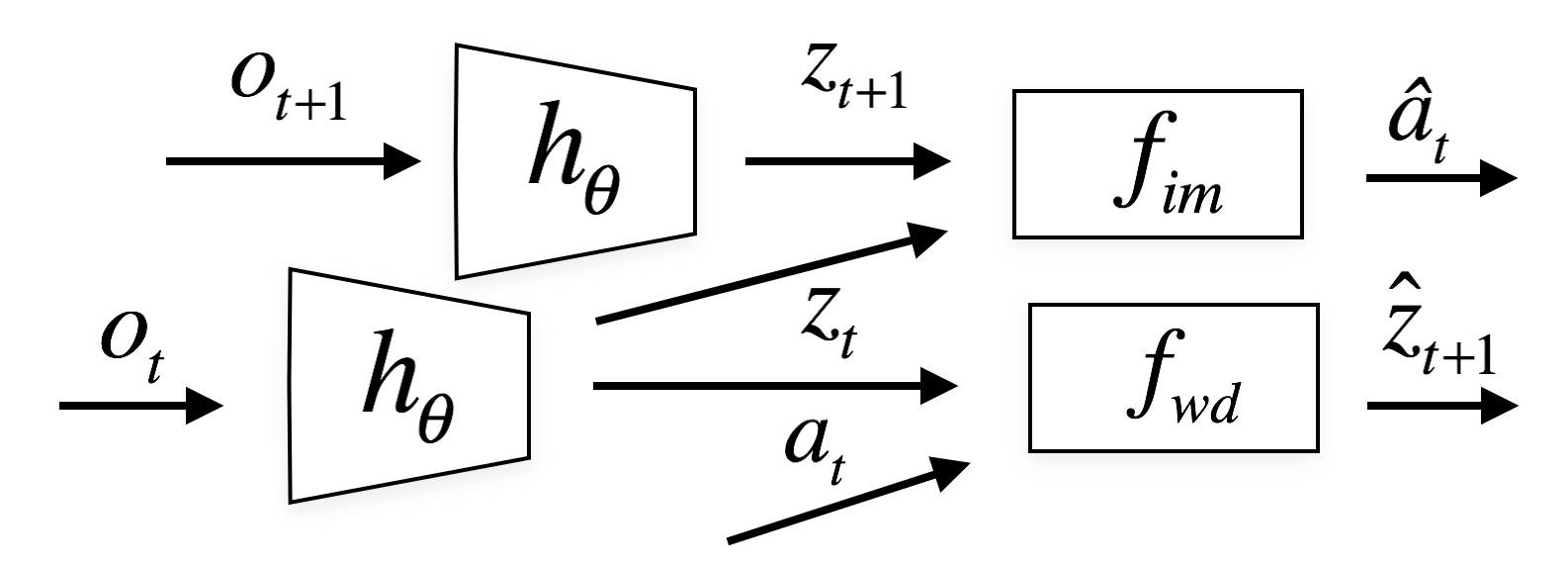}     & \footnotesize{$s_{fm} = s_{im} + ||f_{wd}(z_t,a_t) - z_{t+1}||$} \\
     % \midrule
     % (d) & Regularized BC        & \includegraphics[height=1.6cm]{figs/bc_reg} & \footnotesize{$s_{rbc} =  s_{bc} + s_{im}$} \\
    \bottomrule
    \end{tabular}
    \caption{Proposed action-based supervised 
    % self-supervised
    losses based on expert actions. To train the encoder $h_\theta$, these are minimized over the dataset $\Du$.}
	\label{fig:self_supervised_methods}
  \vspace{-3mm}
\end{figure}

\subsection{Action-based Supervised Stage
% before: Self-supervised
}
\label{ssec:sss}

At the action-based supervised 
% self-supervised
stage the objective is to train an encoder $h_\theta$ to produce a set of features $z_t$ (encoder's bottleneck) given an input observation $o_t$. With this purpose, we have studied some alternatives illustrated in \Fig{self_supervised_methods}. At training time, all of them rely on $\Du$, but they use different inputs and losses to be minimized. We summarize these alternatives in the following.

\mypara{Behavior cloning (BC).} Common deep architectures trained by BC consist of an encoder $h_\theta$ extracting features $z_t$ from observations $o_t$ ({\ie} $z_t=h_\theta(o_t)$), and a fully-connected projection network, $f_{a}(z_t)$, which predicts an expert action $\hat{a}_t$ from such features. In this paper, we follow the architecture presented in \cite{Codevilla:2019}; however, instead of using the high level command ($c_t$) for branching to different projection functions, since our goal is to pre-train a useful representation not performing actual driving with it, we use $c_t$ as part of $o_t$. Therefore, the encoder of \cite{Codevilla:2019} is modified to input the high level command the same way as the speed variable ($v_t$). The input image is processed by a ResNet34 backbone. The following alternatives, also rely on this encoder architecture.

\mypara{Inverse model.} By considering not only $o_t$ but also the subsequent observation $o_{t+1}$ as input, we turn BC into an Inverse model \cite{Agrawal:2016,Shelhamer:2017}. In this case, thinking of the encoder's bottleneck as encoding an agent (driver) internal state, the problem to solve consists of predicting the action that transforms the state $z_t$ into the state $z_{t+1}$. Differently than BC, with an inverse model the actions can come from either an expert driver or just random roaming (or poor driving). 

\mypara{Forward model.} In this case, we want to learn an encoder that is able to output a state $z_t=h_\theta(o_t)$ such that, given an action $a_t$, we can predict the future state as $z_{t+1}=f_{wd}(z_t,a_t)$. However, this can lead to the trivial solution $z_t={\bf 0}$ from which $f_{wd}$ can still produce $z_{t+1}$. Such degenerated encoders $h_\theta$ are of course not of interest. Therefore, we use the regularization strategy of \cite{Agrawal:2016}, consisting of adding also the Inverse model so that the encoded observation $z_t$ is able to predict the action too. Again, the actions can come from either an expert driver or just random roaming. 

%\mypara{Regularized behavior cloning.} Analogously to the previous case, we augment behaviour cloning with the Inverse model, now with the purpose of considering more than one frame for predicting the expert action.

\subsection{Weakly Supervised Stage: Learning Affordances}
\label{ssec:a}
The affordances used in this paper (\Fig{affordances}) consider explicit detection of hazards involving pedestrians and vehicles, respecting traffic lights and considering the heading of the vehicle within the current lane. More specifically, we have considered the following four affordances:

\mypara{Pedestrian hazard} (${hp}_t$). This variable is set to one if there is a pedestrian in our lane at a distance lower than 10 m; otherwise, is set to zero.

\mypara{Vehicle hazard} (${hv}_t$). This variable is set to one if there is a vehicle in our lane at a distance lower than 10 m; otherwise, is set to zero. Vehicles refers to cars, vans, motorbikes, and cyclists.

\mypara{Red traffic light} (${hr}_t$). This variable is set to one if there is a traffic light in red affecting our lane at a distance lower than 10 m; otherwise, is set to zero.

\mypara{Relative heading angle} ($\psi_t$). This variable accounts for the relative angle of the longitudinal vehicle axis with respect to the lane in which it is navigating. The variable runs on $[-\pi,\pi]$~rad with $\psi_t=0$ when vehicle and lane are aligned (no matter the lateral vehicle position within the lane). 

Note that $\{{hp}_t, {hv}_t, {hr}_t\}$ are binary variables, so predicting them is a binary classification problem, while $\psi_t$ is a real number, thus, predicting it is a regression problem. These binary variables are critical to perform stop-\&-go maneuvers by any controller relying on these affordances, while the regressed angle is critical to properly navigating without going out of the lane. Overall, the idea behind these affordances is that relevant visual competences for driving emerge when training the corresponding models; for instance, some kind of pedestrian and vehicle detection, red traffic light detection, and localization of the vehicle within the lane for proper navigation.

%Given the weak supervision in terms of affordances and the pre-trained $h_\theta$, we train a MLP $g_\phi$ as 

The affordance prediction model, $g_\phi$,  that predicts  $\{{hp}_t, {hv}_t, {hr}_t,\psi_t\}$ is obtained by training a MLP, which receives
the output from the pre-trained $h_\theta(o_t)$ as input.% In our study, we have considered the following two cases.

%\mypara{Linear classification/regression.} In this case, inspired by \cite{Alain:2017}, $h_\theta$ is frozen, and  $g_\phi$ is either a linear classifier for each binary-valued affordance, or a linear regressor for the real-valued affordance.   

%\mypara{Fine-tuning.} In this case, $h_\theta(o_t)$ is further processed by a three-layer MLP. This MLP and $h_\theta$ are fine-tuned by using the weakly supervised data. 

%%%%%%%%%%%%%%%%%%%%%%%%%%%%%%%%%%%%%%%%%%%%%%%%%%%%%%%%%%%%%%%%%%%%%%%%%%%%%%%%%%%%%%%%%%%%%%%%%%%%%%%%%%%%%%%%%%%%%%%%%%%%%%%%%%%%%%%%%%%%%%%%%%%%%%%%%%%%%%%%%%%%%%%%%%%%%%%%%%%%%%%%%%%%%%%%%%%%%%%%%%%%%%%%%%%%%%%%%%%%%%%%%%%%%%%%%%%%%%%%%%%%%%%%%%%%

%trim={<left> <lower> <right> <upper>}

\section{Experimental Results}
%\subsection{Implementation Details}

\mypara{Environment.}
As in most recent works addressing autonomous driving, we perform our experiments and data collection in the CARLA simulator \cite{Dosovitskiy:2017}, in particular, using version 0.9.6.  We rely on the widely used Town01 as training and Town02, the new town,
for testing, from now on denoted as T1 and T2. More details about the environment and the benchmarks are provided on the supplementary material.

\mypara{Dataset.} In order to collect the action-based supervised
% Before: self-supervised
dataset ($\Du$) and the weakly supervised dataset ($\Dl$), we modified the default CARLA's autopilot for not only recording $o_t$ and $a_t$, but also our image-level affordances ($y_t$). We collected $\sim50$~hours of image sequences in T1 for training purposes, balancing the training weather conditions, at 20 fps. In this data collection process, we have three cameras, a forward-facing (central) camera from which we will drive at testing time, and two lateral cameras only used for training purposes as in \cite{Bojarski:2016,Codevilla:2018}. Thus, in terms of samples to train $h_\theta$, we have $N_u\sim108,000,000$. This dataset plays the role of $\Du$, while to play the role of $\Dl$ we selected $\Du$'s sub-sequences corresponding to $1\%$ and $10\%$ of the
total amount. In this case, we only consider images acquired by the central camera; thus, totalling $N_l\sim36,000$ for $1\%$ and $N_l\sim360,000$ for $10\%$. These sub-sequences were selected semi-randomly to ensure that jointly form a dataset where the relative heading angle approximates a Gaussian distribution centered at $\psi_t=0$.  Finally, for testing purposes, two new datasets were collected, namely, by driving $\sim1$~hour in T1 and also $\sim1$~hour in T2. This driving was balanced among all weather conditions, and only the central camera is considered; thus, for each town we have $\sim72,000$ images.   

\mypara{Baselines.} In \SSect{sss} we have presented the action-based pre-training strategies for $h_\theta$ that we want to study. In addition, we have incorporated ST-DIM \cite{Anand:2019}, a contrastive representation learning baseline used by agents playing Atari games. We have modified the code provided by the authors just to include ResNet34 as backbone, {\ie} as for the rest of pre-training strategies. In short, ST-DIM is trained to answer if two frames are consecutive or not, without any action-related information involved. Moreover, for the Inverse, Forward, and ST-DIM strategies, we have included seldom variants which require to collect additional $\sim20$~hours of image sequences in T1. However, in this case, instead of relying on our expert driver autopilot, the driving was \emph{random}; thus, eventually running into accidents, driving over the sidewalk, in the wrong lane, etc. In short, navigating by random actions. 
As additional baselines, we have used ImageNet and no pre-training (random initialization). 
%Finally, as upper-bound we have trained the affordance model by just using affordances supervision and ImageNet pre-training. In order to train this model, we have collected new data by running the expert autopilot in T1 for $\sim5$~h, considering only the central camera; thus, $N_l\sim360,000$. 

% Needs to be reaccesed for the new format.
\mypara{Training details.} We train the encoder $h_\theta$ for 100K iterations (mini-batches) using $\Du$. Then, using $\Dl$  we train the affordance prediction model, $g_\phi$, for 20K iterations, no matter if this stage rely on linear classification/regression or fine-tuning. These iteration values were found by preliminary experiments where we monitored training convergence. When $\Dl$ assumes 10\% of
the dataset, we iterated 100K. In all cases, we used ADAM optimizer with an initial learning rate of 0.0002 and batch size of 120. Moreover, for any kind of training, after 75K iterations the learning rate becomes half.  
For obtaining reliable results, we repeat both the encoder and affordance training over three different random seeds, and pick up the model with best performance on training town for driving. As observed in \citep{Codevilla:2019} we saw
a meaningful random seed variation for training the encoders.

\mypara{Controller.} In order to perform driving evaluations, we have tuned a PID controller that takes the estimated affordances as input and outputs the action commands ($a_t$) to control the AV. Given perfect affordances ({\ie} those annotated as ground truth), this controller has been tuned to drive well in T1 (lateral and longitudinal control). 

%%%
% This are the evaluation stratatgies
%%

\subsection{Results}
\label{sec:er}

%\subsection{Experimental settings}

\begin{comment}
\begin{table*}[t!]
\footnotesize
\centering
\resizebox{0.90\linewidth}{!}{
\begin{tabular}{@{}lccccc|cccc@{}}
\toprule
 \textbf{Pre-training} && Pedestrian ($hp$) & Vehicle ($hv$) & Red T.L. ($hr$)  && Left Turn & Straight & Right Turn  \\
\midrule
Random &&  $26\pm0$ &  $50\pm1$ &  $42\pm0$ &&  $4\pm1$  & $40\pm1$   & $1\pm0$\\
ImageNet && $37\pm2$ & $75\pm0$ &  $47\pm0$ && $5\pm2$ & $27\pm1$ &  $3\pm0$ \\
\midrule
Contrastive (ST-DIM)  && $47\pm1$ & $53\pm0$ & $53\pm0$ && $13\pm1$  & $30\pm1$  & $3\pm0$ \\
\midrule
Forward  && $50\pm0$ & $63\pm0$ & $60\pm0$ && $43\pm0$ &  $90\pm0$  & $66\pm0$ \\
Inverse  && $49\pm0$ & $78\pm0$ & $70\pm0$ && $\textbf{67}\pm0$ & $\textbf{93}\pm1$  & $\textbf{85}\pm0$ \\
Behavior Cloning (BC) && $47\pm0$ & $\textbf{81}\pm0$ & $\textbf{75}\pm0$ && $49\pm1$ & $68\pm1$ & $70\pm1$ \\
%Regularized BC (RBC)  &&  $\textbf{58}\pm0$ & $69\pm0$ &  $74\pm0$ && $61\pm3$ & $62\pm0$ & $75\pm1$ \\
%\midrule
%ImageNet$\!+\!\!$ Supervised (5h) && $73\pm1$ & $95\pm1$ & $81\pm1$ && $75\pm0$ & $90\pm2$ & $86\pm1$ \\
\bottomrule
\end{tabular}
}
\caption{Linear probing results. Left: F1 score  for the binary affordances. Right: threshold relative error (TRE) for the relative angle, divided depending on driving situation. For all training strategies, we have performed the full training protocol three times (F1-score or TRE) to obtain the mean and standard deviation. Results scaled by 100 for visualization purposes.}
\label{tbl:0.25h_frozen}
\end{table*}
\end{comment}

\begin{table*}[t!]
\footnotesize
\centering
\resizebox{0.95\linewidth}{!}{
\begin{tabular}{@{}lccccccccc@{}}
\toprule
&& \multicolumn{3}{c}{Binary Affordances} && \multicolumn{3}{c}{Relative Angle ($\psi_t$)} \\
 \textbf{Pre-training} && Pedestrian ($hp$) & Vehicle ($hv$) & Red T.L. ($hr$)  && Left Turn & Straight & Right Turn  \\
\midrule
No pre-training &&  $26\pm0$ &  $50\pm1$ &  $42\pm0$ &&  $11.38\pm0.18$  & $1.85\pm0.03$   & $24.68\pm0.03$\\
ImageNet && $37\pm2$ & $75\pm0$ &  $47\pm0$ && $11.69\pm0.57$ & $2.83\pm0.07$ &  $25.55\pm0.10$ \\
\midrule
Contrastive (ST-DIM)  && $47\pm1$ & $53\pm0$ & $53\pm0$ && $10.43\pm0.21$  & $2.62\pm0.03$  & $18.75\pm0.23$ \\
\midrule
Forward  && $\textbf{50}\pm\textbf{0}$ & $63\pm0$ & $60\pm0$ && $5.35\pm0.03$ &  $0.52\pm0.00$  & $6.61\pm0.03$ \\
Inverse  && $49\pm0$ & $78\pm0$ & $70\pm0$ && $\textbf{3.57}\pm\textbf{0.03}$ & $\textbf{0.46}\pm\textbf{0.00}$  & $\textbf{3.78}\pm\textbf{0.06}$ \\
Behavior Cloning (BC) && $47\pm0$ & $\textbf{81}\pm\textbf{0}$ & $\textbf{75}\pm\textbf{0}$ && $4.89\pm0.03$ & $1.24\pm0.03$ & $6.25\pm0.10$ \\
\bottomrule
\end{tabular}
}
\caption{Linear probing results. Left: F1 score for the binary affordances (higher is better). Results are scaled by 100 for visualization purpose. Right: MAE of the relative angle (lower is better), shown for different navigation maneuvers. MAE is shown in degrees for an easier understanding. 
%For all training strategies, we have performed the full training protocol three times (F1-score or MAE), thus, we report the mean and standard deviation for each case.
}
\label{tbl:0.25h_frozen}
\end{table*}

% First paragraph experimental settings

\begin{figure*}[t!]
\centering
\includegraphics[width=0.9\linewidth]{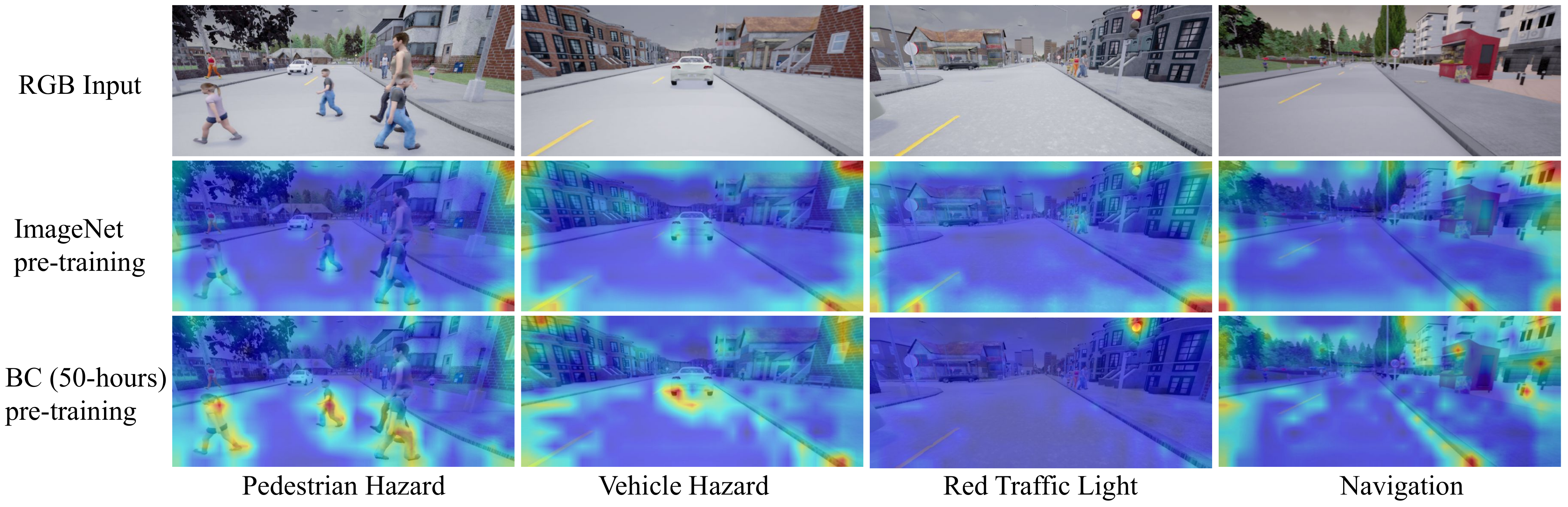}
\caption{\textbf{Attention heatmaps.} Top: RGB input images from a town unseen during training. Mid: attention heatmaps of ImageNet pre-trained encoder. Bottom: attention heatmaps of a BC encoder pre-trained with 50 hours of expert driving data. From left to right, we show cases involving different affordances: pedestrian hazard, vehicle hazard, red traffic light detection and navigation.}
\label{fig:visualresults}
\end{figure*}

\mypara{Linear probing.} We start by evaluating the representation learning capabilities of action-based methods using the commonly applied linear probing technique \cite{Alain:2017,Oord:2018,Chen:2020,He:2019}.
More specifically, using a frozen $h_\theta$ as feature extractor, we train a linear classifier to predict affordances with an affordance dataset $\Dl = 1\%$. %The training supervision is provided by assuming that the 1\% of the unlabelled images used to train $h_\theta$ are labelled with their corresponding affordances. %\alertAnt{How the 1\% is obtained, randomly? with some restricitons?}  
Each trained model is tested in the $\sim1$~hour testing set for T2, not seen during training. In order to assess the performance of binary-affordance models we use F1-score, while for assessing the performance of the relative heading angle, we use MAE. For better analysis, we divide the relative heading angle into three cases, left turn, straight and right turn, according to the navigation situation.  We consider as left regime those cases where the relative heading angle ground truth is lower than $-0.1$ rad, as right regime when it is larger than $0.1$ rad, and as straight regime otherwise. For each pre-training strategy, we repeat linear classifier training with three random seeds, and compute its mean and standard deviation in Tables \ref{tbl:0.25h_frozen} and \ref{tbl:0.25h_frozen_random}.
 
 %the threshold relative error (TRE) since it is more correlated to driving than MSE \cite{Codevilla:2018b} and provides a more interpretable result.  % MIGHT NOT BE TRE EXACTLY
 %In summary the TRE reports the percentage of time your regressor is able to predict a the relative angle with an error within an threshold. We also discriminate results according to the lane appearance by reporting the TRE according to three regimes, namely, straight segment, left and right curves.

% Second paragraph explanation
Table \ref{tbl:0.25h_frozen} shows the F1/MAE scores for
the affordances prediction. Note that those results consist of zero-shot generalization to an unseen town (T2). %Further results, also with T1, are presented on the supplementary material.
The main observation is that action-based pre-training (Forward/Inverse/BC)  outperforms all the other reported pre-training strategies.
%\alertAnt{This particular example is not useful, it just compares pre-training vs no pre-training, is there any particular reason for this choice?}{The difference is particularly  expressive when comparing BC and no pre-training (Random) where the vehicle affordance, for instance, can be much more reliably inferred.}
However, we see that the action-based pre-training is mostly beneficial to help on reliably estimating the vehicle's relative angle with respect to the road. The constrastive method, ST-DIM, shows promising results for the binary affordances but results on very poorly relative angle estimations. We also observed a poor generalization capability for ImageNet pre-training. These results suggest that a useful scene representation is learned by training encoders with expert demonstration data. Additional evidence is presented in \Fig{visualresults}, which shows examples of attention heatmaps from an ImageNet encoder, and a BC encoder that was trained with 50 hours of data. These attention maps are calculated by the simple average of the feature maps from the third ResNet34's block. We can see that the necessity to imitate expert demonstrations creates activations on useful objects such as pedestrians, vehicles, traffic lights and lane markings; which is in agreement with the fact that a linear classifier can predict well the set of affordances with the action-based pre-training. 
 
We also further study if the source data needs to come from expert driving or from random actions
as in \cite{Agrawal:2016}. 
The general intuition is that
 the Inverse model can learn a good representation by
 learning the dynamics of the scene. We show on Table \ref{tbl:0.25h_frozen_random} that inverse model can,
 indeed, outperform the no pre-training condition even when
 using random actions. % compares
%the results of using different action-based techniques that rely on data from a random policy demonstrator. % Describe this 
However, we show that there is a lot more benefit for representation learning obtained from the expert action information than from random action. For ST-DIM, as expected, the difference between random and expert policy is smaller since it is not
based on action.
% Yet, random policy can yield promising results, specially for contrastive methods, \alertAnt{This is only true for ST-DIM and this performs bad for regression affordances}{}
%Also, some random action results, specially on the relative angle prediction are still able to outperform the image-net pre-training.

\begin{table*}[!]
\footnotesize
\centering
\resizebox{0.95\linewidth}{!}{
\begin{tabular}{@{}lccccccccc@{}}
\toprule
&& \multicolumn{3}{c}{Binary Affordances} && \multicolumn{3}{c}{Relative Angle ($\psi_t$)} \\
 \textbf{Pre-training} && Pedestrian ($hp$) & Vehicle ($hv$) & Red T.L. ($hr$)  && Left Turn & Straight & Right Turn  \\

\midrule
No pre-training &&  $26\pm0$ &  $50\pm1$ &  $42\pm0$ &&  $11.38\pm0.18$  & $1.85\pm0.03$   & $24.68\pm0.03$\\
\midrule
Contrastive (ST-DIM)  && $41\pm0$ & $62\pm1$ & $63\pm1$ && $9.01\pm0.46$  & $2.77\pm0.18$  & $18.37\pm0.45$ \\
Contrastive Random (ST-DIM)  && $39\pm1$ & $\textbf{73}\pm\textbf{1}$ & $47\pm0$  && $9.70\pm0.41$ &  $2.98\pm0.11$ & $15.89\pm0.41$\\
\midrule
Forward  && $\textbf{50}\pm\textbf{0}$ & $51\pm0$ & $58\pm0$ && $4.87\pm0.00$ &  $0.52\pm0.00$  & $6.07\pm0.06$ \\
Forward Random  && $20\pm1$ & $38\pm0$ & $16\pm0$ && $11.54\pm0.03$ & $1.20\pm0.00$ & $19.14\pm0.00$  \\
\midrule
Inverse  && $45\pm0$ & $66\pm0$ & $\textbf{73}\pm\textbf{0}$ && $\textbf{3.02}\pm\textbf{0.03}$ & $\textbf{0.42}\pm\textbf{0.03}$  & $\textbf{5.06}\pm\textbf{0.17}$ \\
Inverse Random && $26\pm0$ & $49\pm0$ & $59\pm0$  && $8.50\pm0.53$ & $1.45\pm0.03$ & $13.14\pm0.34$ \\

\bottomrule
\end{tabular}
}
\caption{Linear probing results comparing encoders trained with random policy training data versus expert demonstration data. Note that, to provide fair comparison, the encoders here were trained with 20 hours data, which are different than the ones from Table \ref{tbl:0.25h_frozen}.}
\label{tbl:0.25h_frozen_random}
\end{table*}

\mypara{Driving results.} We evaluate the driving performance
of our method on the CARLA NoCrash benchmark \cite{Codevilla:2019}, which mainly focuses on the capabilities of models to drive under the presence of pedestrians and vehicles. The objective is
to complete a set of goal-oriented episodes without crashing. The CARLA simulator provides high 
% Traffic light as separate table ?
level planning commands to navigate towards a targeted town location (goal) from current one; these commands are part of the observations used to train $h_\theta$. 
%\alertAnt{Do you refer to high-level navigation planing?}{} \alertYi{Yes, we take the command as a part of the input to train the encoder. We have explained this in Sec.\ref{ssec:sss} (BC).} 
For this evaluation, we updated the CARLA NoCrash benchmark to CARLA 0.9.6 version augmented with the new pedestrian crossing algorithms (recently incorporated to last CARLA version). 
For driving, we fine tune the whole network using three layers as the projection $g_\phi$. We report the success rate (higher is better) on the driving tasks and the percentage of traffic lights crossed in red (lower is better). For each model, we repeat driving for three times, and compute its mean and standard deviation in Tables \ref{tbl:driving_results} and \ref{tbl:driving_results_hours}.

\begin{comment} %% Tabke 3 customized for generating a pdf for my presentation, comment for regular paper
\begin{table*}[t!]
\footnotesize
\centering
\resizebox{0.9\linewidth}{!}{
\begin{tabular}{@{}lccccccccccc@{}}
\toprule
 %\multicolumn{2}{c}{($D^u$=50H.)} && \multicolumn{8}{c}{($D^l$=5H.)} \\
   && \multicolumn{4}{c}{Training Town} && \multicolumn{4}{c}{New Town} \\
    Technique  && Empty & Regular & Dense &  T.L. && Empty & Regular & Dense &  T.L.   \\
  %\midrule
  \toprule
  No pre-training && $78\pm4$ & $79\pm7$ & $48\pm4$ & $12\pm1$ && $57\pm1$ & $37\pm4$ & $10\pm1$ & $18\pm2$ \\
  ImageNet && $86\pm1$ & $\underline{89\pm3}$ & \underline{$66\pm1$} & $12\pm0$ && $21\pm3$ & $19\pm3$ & $13\pm5$ & $23\pm2$ \\
  Behavior Cloning (BC) && $\textbf{91}\pm1$ & $\underline{\textbf{91}\pm4}$ & $\underline{\textbf{68}\pm5}$ & $\textbf{8}\pm1$ && $\textbf{83}\pm3$ & $\textbf{61}\pm4$ & $\textbf{25}\pm4$ & $\textbf{8}\pm1$  \\
\bottomrule
 \end{tabular}
}
\caption{Comparison of action-based pre-training with baselines (top) and other methods from the literature (bottom). We are able to surpass ImageNet pre-training and the CILRS baseline. Results are from the CARLA 0.9.6 NoCrash benchmark.} %\alertYi{We show in bold the highest means, and underline similar results considering standard deviations.}}
\label{tbl:driving_results}
\end{table*}
\end{comment}

\begin{table*}[t!]
\footnotesize
\centering
\resizebox{0.9\linewidth}{!}{
\begin{tabular}{@{}lccccccccccc@{}}
\toprule
 %\multicolumn{2}{c}{($D^u$=50H.)} && \multicolumn{8}{c}{($D^l$=5H.)} \\
   && \multicolumn{4}{c}{Training Town} && \multicolumn{4}{c}{New Town} \\
    Technique  && Empty & Regular & Dense &  T.L. && Empty & Regular & Dense &  T.L.   \\
  %\midrule
  \toprule
  No pre-training && $78\pm4$ & $79\pm7$ & $48\pm4$ & $12\pm1$ && $57\pm1$ & $37\pm4$ & $10\pm1$ & $18\pm2$ \\
  Image Net && $86\pm1$ & $\underline{89\pm3}$ & $66\pm1$ & $12\pm0$ && $21\pm3$ & $19\pm3$ & $13\pm5$ & $23\pm2$ \\
  \midrule
  Contrastive (ST-DIM) && $73\pm2$ & $84\pm4$ & $\underline{\textbf{71}\pm3}$ & $10\pm1$ && $66\pm5$ & $49\pm2$ & $17\pm0$ & $21\pm1$  \\
  \midrule
  Forward && $68\pm3$ & $84\pm3$ & $\underline{68\pm8}$ & $\underline{9\pm1}$ && $49\pm5$ & $37\pm3$ & $18\pm5$ & $13\pm2$  \\
  Inverse && $73\pm4$ & $82\pm2$ & $61\pm3$ & $\underline{\textbf{8}\pm1}$ && $\underline{\textbf{83}\pm2}$ & $\textbf{67}\pm8$ & $\underline{\textbf{26}\pm8}$ & $\underline{\textbf{8}\pm0}$  \\
  Behavior Cloning (BC) && $\textbf{91}\pm1$ & $\underline{\textbf{91}\pm4}$ & $\underline{68\pm5}$ & $\underline{\textbf{8}\pm1}$ && $\underline{83\pm3}$ & $61\pm4$ & $\underline{25\pm4}$ & $\underline{\textbf{8}\pm1}$  \\
  %Regularized BC (RBC) && $82\pm1$ & $89\pm6$ & $68\pm0$ & $9\pm1$ && $\textbf{83}\pm5$ & $62\pm3$ & $24\pm5$ & $18\pm1$  \\
%\midrule
% ImageNet$\!+\!\!$ Supervised (6h) && 90 & 90 & 51 &&& 16 &&  \\
%\midrule
\midrule
 CILRS 0.8.4 \cite{Codevilla:2019} && $97\pm2$ & $83\pm0$ & $42\pm2$ & $47$ && $66\pm2$ & $49\pm5$ & $23\pm1$ & $64$ \\
 %CIRLS 0.9.6 && $72\pm5$ & $47\pm2$ & $20\pm3$ & $74\pm2$ && $71\pm2$ & $42\pm7$ & $12\pm3$ & $63\pm1$ \\
  LBC \cite{Chen:2019} && $97\pm1$ & $93\pm1$  & $71\pm5$ & N/A && $100\pm0$ & $94\pm3$ & $51\pm3$ & N/A  \\
\bottomrule
 \end{tabular}
}
\caption{Comparison of action-based pre-training with baselines (top) and other methods from the literature (bottom). We are able to surpass ImageNet pre-training and the CILRS baseline. Results are from the CARLA 0.9.6 NoCrash benchmark.} %\alertYi{We show in bold the highest means, and underline similar results considering standard deviations.}}
\label{tbl:driving_results}
\end{table*}

Table \ref{tbl:driving_results} compares the performance 
of the action-based methods with the baselines and other
methods from the literature. For all implemented
methods, we considered fine-tuning with $\Dl$ = $10\%$.
% Comparing itself
Firstly, we see that Inverse model and BC % \alertAnt{I have corrected in some places, but there is not a consistent terminology, somotimes BC, sometimes Behaviour Cloning, sometimes Inverse Model, sometimes inverse model, sometimes ST-DIM, sometimes STDIM, action based vs action-based, etc.}{} 
are the best representation learning strategies to pre-train the encoder, specially in new town.
%The forward model obtained clear inferior performance on generalization while we inverse model had marginally better results.% DO we readd RBC ?
% Comparing with STDIM
Both models also clearly outperform the constrastive-based baseline (ST-DIM) in new town. However, in the training town the contrastive method obtained relevant results, specially under dense traffic. 
As reference, we report results from Learning by Cheating (LBC) \cite{Chen:2019} and the CILRS method \cite{Codevilla:2019}, presented on the bottom of Table \ref{tbl:driving_results}. 
%\alertAnt{But with this 0.9.6 and 0.9.7 thing, this may be tricky to explain. Let's think about it.}{Shown results are copied from the corresponding papers.} 
Shown results are copied from the corresponding papers. Our proposed approaches also achieved very close results to the LBC method \cite{Chen:2019} and outperforms CILRS specially when reacting to traffic lights.  Note that the LBC method requires high supervision for training a teacher network, which teaches a student network to drive end-to-end. Our method uses much less densely annotated data and does not use dataset aggregation (Dagger \cite{ross2011reduction}). Moreover, note that reported results from CILRS are on CARLA version 0.8.4, the benchmark from the newer version is considerably more difficult. We compare both on the supplementary material.

\begin{table*}[t!]
\footnotesize
\centering
\resizebox{0.9\linewidth}{!}{
\begin{tabular}{@{}lcclcccccccccc@{}}
\toprule
 && && \multicolumn{4}{c}{Training Town} && \multicolumn{4}{c}{New Town} \\
  ($D^l$) &&  ($D^u$)   && Empty & Regular & Dense &  T.L. && Empty & Regular & Dense &  T.L.   \\
\midrule
 \multirow{4}{*}{0.5 hours}  && No Data && $25\pm4$ & $11\pm3$ & $3\pm2$ & $51\pm2$ && $1\pm1$ & $1\pm1$ & $0\pm0$ & $57\pm4$   \\
 && 5 hours && $68\pm5$ & $58\pm5$ & $27\pm5$ & $\underline{\textbf{10}\pm1}$ && $14\pm1$ & $7\pm1$ & $3\pm1$ & $23\pm2$   \\
 && 20 hours && $\textbf{98}\pm1$ & $\textbf{93}\pm2$ & $\textbf{61}\pm6$ & $\underline{\textbf{10}\pm1}$ && $17\pm2$ & $12\pm2$ & $2\pm2$ & $15\pm2$   \\
 && 50 hours && $95\pm2$ & $87\pm2$ & $47\pm2$ & $12\pm1$ && $\textbf{25}\pm3$ & $\textbf{20}\pm4$ & $\textbf{6}\pm1$ & $\textbf{10}\pm1$ \\
\midrule
 \multirow{4}{*}{5 hours} && No Data && $78\pm4$ & $79\pm7$ & $48\pm4$ & $12\pm1$ && $57\pm1$ & $37\pm4$ & $10\pm1$ & $18\pm2$   \\
 && 5 hours && $58\pm6$ & $80\pm4$ & $\underline{69\pm5}$ & $\underline{9\pm1}$ && $68\pm6$ & $55\pm5$ & $23\pm3$ & $12\pm2$   \\
 && 20 hours && $74\pm2$ & $81\pm4$ & $\underline{\textbf{70}\pm5}$ & $10\pm1$ && $\underline{82\pm2}$ & $\underline{\textbf{66}\pm5}$ & $\textbf{34}\pm0$ & $12\pm1$   \\
 && 50 hours && $\textbf{91}\pm1$ & $\textbf{91}\pm4$ & $\underline{68\pm5}$ & $\underline{\textbf{8}\pm\textbf{1}}$ && $\underline{\textbf{83}\pm3}$ & $\underline{61\pm4}$ & $25\pm4$ & $\textbf{8}\pm1$  \\
\bottomrule
 \end{tabular}
}
\caption{Driving performance ablation study of the BC pre-training encoder investigating the quantity of expert driving data ($D^u$) given a smaller amount of annotated affordances ($D^l$).}% \alertAnt{H. should be either hours or h for coherence with the rest of the paper}{}} %\alertAnt{Results from the CARLA 0.9.6 NoCrash benchmark. // This is also in Table 3, right?}}{}
\label{tbl:driving_results_hours}
\end{table*}

%%% Driving
\mypara{Ablation study.} In Table \ref{tbl:driving_results_hours}, we analyse the data amount impact of both supervised data and expert demonstrations. We used the simplest pre-training encoding, BC. We observe a clear correlation between performance improvement and the quantity of action-based supervised data $D^u$. We can see that with only 30 minutes (0.5 hours) of annotated data, the pre-trained network has a performance close to our best reported results in training town. For the generalization conditions, however, only 0.5 hours of annotated data is not enough to obtain satisfactory results. Pre-training is useful also when using higher amounts of annotated data (5 hours),  as shown on the bottom of Table \ref{tbl:driving_results_hours}. %This improvement is also substantial when the number of unsupervised data is small, for instance, when $D^u = 5 hours$ and $D^l = 5 hours$ .
However, the impact of pre-training when more annotated data is available is clearer in New Town. 
%However, without enough
%supervised affordances data the models failed to generalize.
Note that when comparing 50 hours with 20 hours, on pre-training, the results are similar, since 20 hours driving seems to be sufficient to capture the inherent variability of T1.

\section{Conclusions}
\label{sec:c}

In this paper, we have shown that representations learned by using action-based methods (BC or Inverse model) are promising as pre-trained representations for autonomous driving controllers based on affordances. Moreover, we have found that most of the benefit comes when the driving experiences (actions) are captured from proper driving (humans). In other words, expert driving outperforms random roaming for representation learning. While considerable future research is needed to improve the raw performance of the methods explored here, the fact that the required data can be easily obtained by simply recording the actions of good drivers, highlights the potential of action-based methods for learning representations for autonomous vehicles beyond pure end-to-end autonomous driving models. 

It would be relevant to explore if the pre-training strategy presented here can be also helpful for training annotation-intensive visual models, such as those for semantic class/instance segmentation or object detection. Further, it would be interesting to examine other strategies to use the expert driver data in order to  further improve performance.

%===============================================================================

% The maximum paper length is 8 pages excluding references and acknowledgements, and 10 pages including references and acknowledgements

\clearpage
% The acknowledgments are automatically included only in the final version of the paper.
\acknowledgments{Yi Xiao and Antonio M. López acknowledge the financial support received for this work from the Spanish TIN2017-88709-R (MINECO/AEI/FEDER, UE) project. Antonio acknowledges the financial support to his general research activities given by ICREA under the ICREA Academia Program. Yi Xiao acknowledges the Chinese Scholarship Council (CSC) grant No.201808390010. Yi Xiao and Antonio M. López acknowledge the support of the Generalitat de Catalunya CERCA Program as well as its ACCIO agency to CVC's general activities. Felipe Codevilla and Christopher Pal
would like to thank the Natural Sciences and Engineering Research Council and the Programme d'innovation en Cybersécurité du Quebec for their support.}

%===============================================================================

\bibliography{example}  % .bib

\end{document}

\pagestyle{headings}
%\mainmatter
%\def\ECCVSubNumber{4420}  % Insert your submission number here

\title{Supplementary Material for ``Action-based Representation Learning for Autonomous Driving ''} % Replace with your title

%\begin{comment}
%\titlerunning{ECCV-20 submission ID \ECCVSubNumber} 
%\authorrunning{ECCV-20 submission ID \ECCVSubNumber} 
%\author{Anonymous ECCV submission}
%\institute{Paper ID \ECCVSubNumber}

\maketitle

% Solving problems on the real world requires quick generalization to new situations and the ability to react
% to certain situations under a low amount of data.
% DNN replaced the use of handcraft features with general
%learned features from a huge dataset such ImageNet that  has allowed hundreds of problems to be solved on a low data regime.

\section{NoCrash Benchmark: CARLA 0.9.6 (modified)}
For performing the driving evaluation, we updated the \textit{NoCrash} benchmark to work with version 0.9.6.
This version has some differences compared with CARLA 0.8.4, the main one is related to pedestrians that now have a very different crossing pattern. In addition, they are now able to cross roads in groups and are more equally spread over the town.
Thus, it is necessary to change the number of pedestrians spawned on the town in order to reproduce the benchmark from version 0.8.4. For the regular task, we increased the number of pedestrians from 50 to 125 for Town01, and 50 to 100 for Town02. In
dense task, we increased
from 250 to 400 for Town01, and  from 150 to 300 for Town02.

The routes have been also changed since the new API allowed a much
more controlled sampling of the routes to be used on the benchmark. Thus, we made a few
changes in the routes to guarantee they followed the  restrictions described on \cite{Dosovitskiy:2017}. We restrict at least 1000 meters distance between start and end point for Town01, and 500 for Town02.

\begin{figure}
	\centering
    \begin{tabular}{c@{\hspace{5mm}}c@{\hspace{5mm}}c}
	    \includegraphics[height=1.66cm]{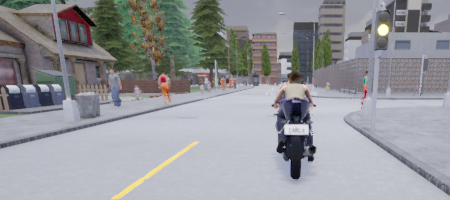} &
      \includegraphics[height=1.66cm]{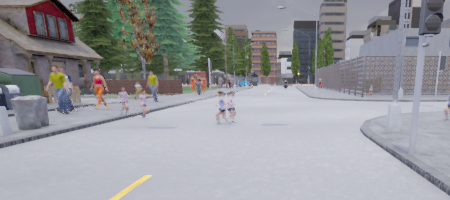} &
      \includegraphics[height=1.66cm]{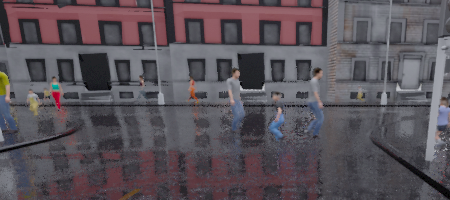}
      % MAYBE MAKE IT IS SMALLER

       %\multicolumn{2}{c}{\includegraphics[height=1.5cm]{figs/control}} \\
       %\multicolumn{2}{c}{(c) tune controller}
       
    \end{tabular}
    \caption{New types of vehicles and pedestrians present on the updated version of the benchmark. Left: motorbikes were not included on the previous benchmark. Middle: kids are now also part of pedestrians. Right: now pedestrians are able to better agglomerate when crossing roads.}
	\label{fig:cases}
  \vspace{-3mm}
\end{figure}

The visuals also have been slightly
changed especially with respect
to the dynamic obstacles.
We can see on Figure \ref{fig:cases} some new features of the benchmark that are present
on version 0.9.6.
The vehicles now include bikes, motorbikes and
infant pedestrians.
Our update was analogous to the one done by \cite{Chen:2019}, however,
instead of implementing it ourselves,
we  added to version 0.9.6 the pedestrian navigation algorithm provided by the official CARLA 0.9.7 version. Note that for this paper, the version used was still CARLA 0.9.6, while only the pedestrian navigation from 0.9.7 was added. We decided to stay on version 0.9.6 since newer versions of CARLA have incorporated major differences on traffic management that drastically changed vehicle behavior. 

\section{Data distribution}

\begin{figure}[h!]
 \centering
  \includegraphics[width=0.48\linewidth]{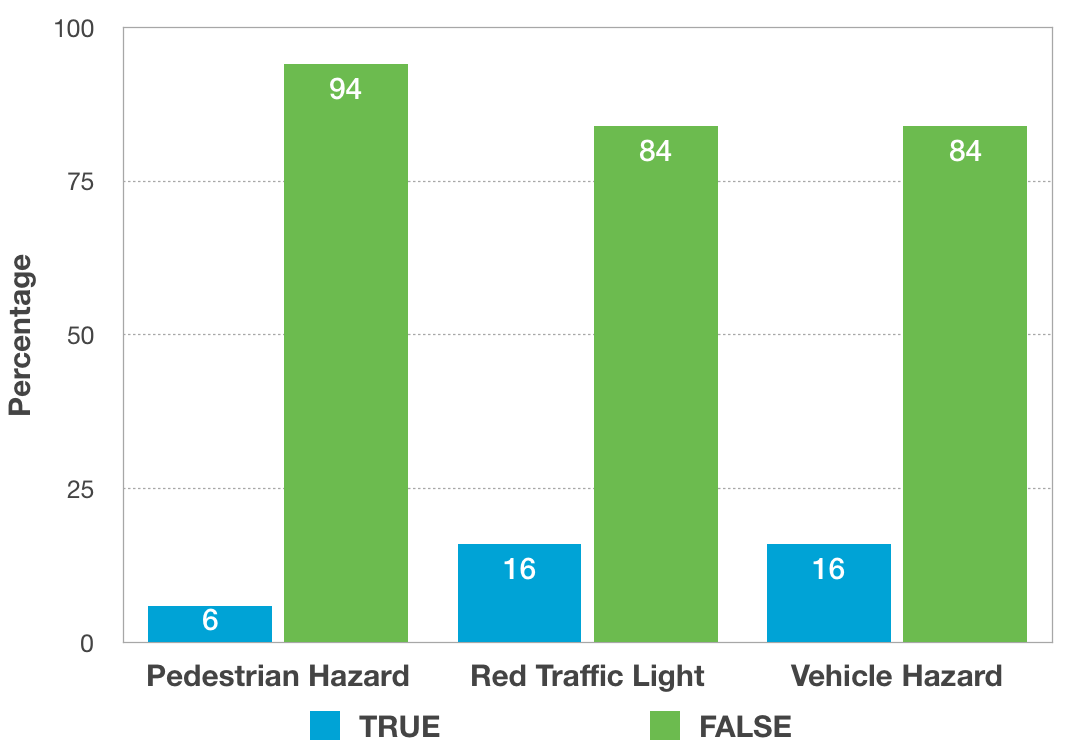}
  \includegraphics[trim={0 0 0 8mm}, clip,width=0.46\linewidth]{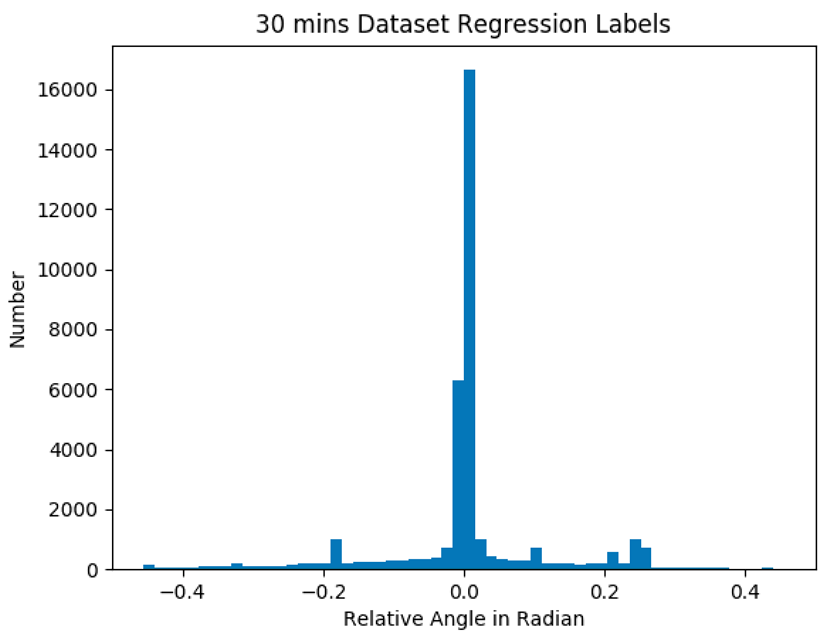}
  \caption{The distributions of affordances on the $\sim30$ minutes training dataset.}
  \label{fig:classification_labels}
\end{figure}

In Figure \ref{fig:classification_labels}, we show the distributions of the 30 minutes annotated dataset, which was used for training the affordances prediction network $g_\phi$. This dataset is a subset of the full 50 hours dataset, carefully sampled to maintain the same distribution as the full dataset.
%we show the distributions of the 30 minutes data， excerpt used for training the affordances prediction network $g_\phi$. We sample this excerpt based on the distribution of the full 50 hours dataset.

\section{Network Architectures}

\begin{figure}[h!]
 \centering
  \includegraphics[width=0.6\linewidth]{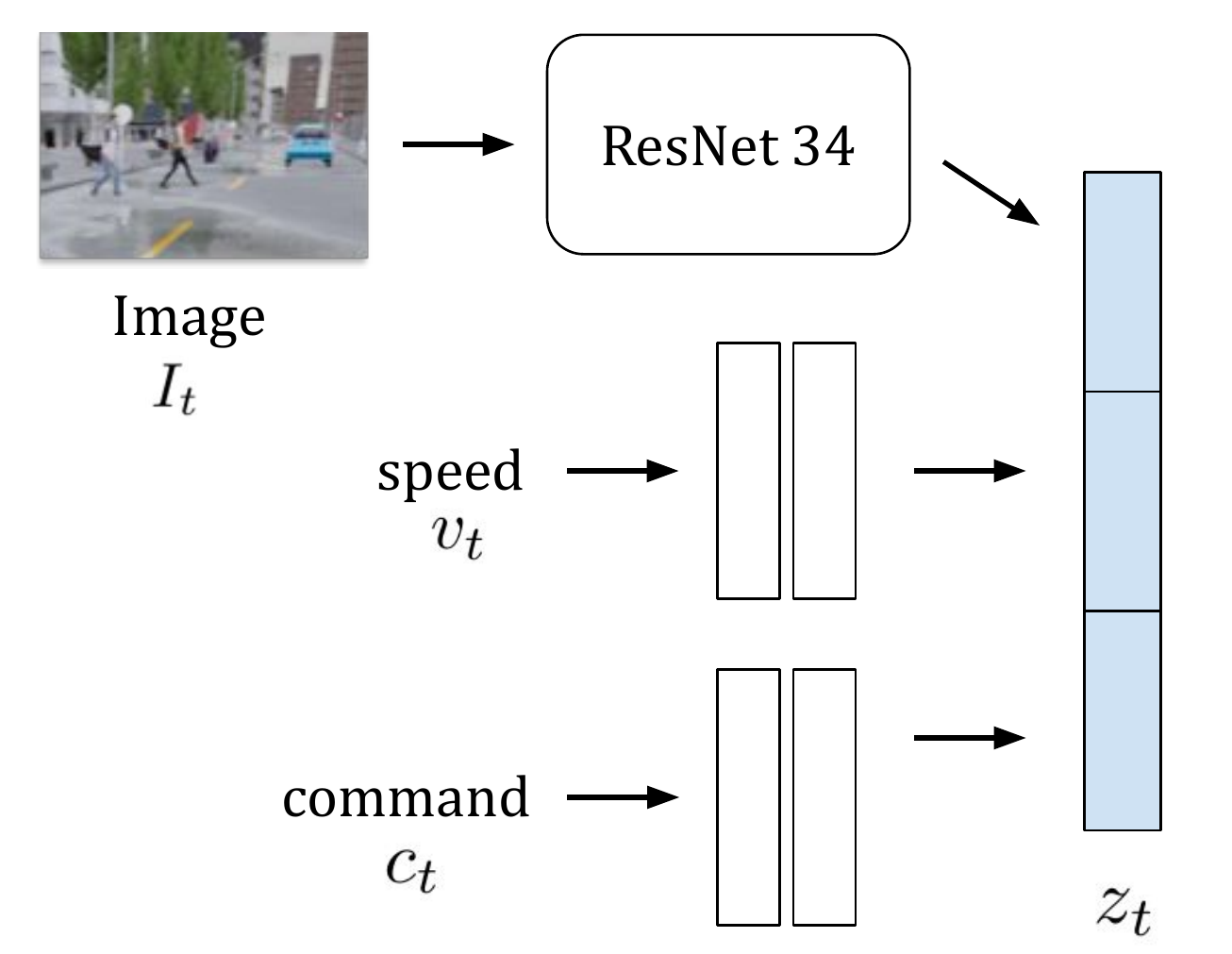}
  \caption{The architecture of the encoder $h_\theta$. Blue rectangles indicate that the features are appended. White rectangles are fully connected layers.}
  \label{fig:encoder}
\end{figure}

In Figure \ref{fig:encoder}, we detail the architecture of our encoder, $h_\phi$. We follow the architecture presented in \cite{Codevilla:2019} using the high level command,  $c_t$, the image input, $I_t$ , and the speed variable  $v_t$, as part of $o_t$. 
In Table \ref{tab:BC_configuration}, we detail the parameters of network architectures for different action-based
representation learning approaches. We also detail the affordance projection network.

\begin{table}[h!]
\begin{center}

\resizebox{0.95\columnwidth}{!}{
\begin{tabular}{cc}
\resizebox{0.5\columnwidth}{!}{
\begin{tabular}{cccc}
\toprule
& Input & Output & Num. of  \\
Module & Dimension & Channels  & Dropout\\
\noalign{\smallskip}
\midrule
ResNet 34 & 200 $\times$ 88 $\times$ 3 & 512 & 0.0\\
\midrule
Speed & 1 & 128  & 0.0 \\
        & 128 & 128 & 0.0 \\
\midrule
Command & 4 & 128  & 0.0 \\
        & 128 & 128 & 0.0 \\
\midrule
 Join & 512+128+128  & 512 & 0.0 \\
\midrule
              & 512  & 256 & 0.0\\
Action Branch & 256  & 256 & 0.5\\
              & 256  & 3 & 0.0\\
\midrule
        & 512  & 256 & 0.0\\
Speed Branch & 256  & 256 & 0.5\\
        & 256  & 1  & 0.0\\ 
\bottomrule
\end{tabular} 
}  &

\resizebox{0.55\columnwidth}{!}{
\begin{tabular}{cccc}
\toprule
& Input & Output & Num. of  \\
Module & Dimension & Channels  & Dropout\\
\noalign{\smallskip}
\midrule
ResNet 34 & 200 $\times$ 88 $\times$ 3 & 512 & 0.0\\
\midrule
Speed  & 1 & 128  & 0.0 \\
        & 128 & 128 & 0.0 \\
\midrule
Command  & 4 & 128  & 0.0 \\
        & 128 & 128 & 0.0 \\
\midrule
 Join & 512+128+128  & 512 & 0.0 \\
 
\midrule
 Join ($Z_t$,$Z_{t+1}$) & 512 + 512  & 512 & 0.0 \\
\midrule
       & 512  & 256 & 0.0\\
Action ($a_t$) & 256  & 256 & 0.0\\
       & 256  & 3 & 0.0\\

\bottomrule
\end{tabular}} \\
 & \\
(a) Behavior Cloning (BC) & (b) Inverse model \\
& \\

\resizebox{0.5\columnwidth}{!}{
\begin{tabular}{cccc}
\toprule
& Input & Output & Num. of  \\
Module & Dimension & Channels  & Dropout\\
\noalign{\smallskip}
\midrule
ResNet 34 & 200 $\times$ 88 $\times$ 3 & 512 & 0.0\\
\midrule
Speed  & 1 & 128  & 0.0 \\
        & 128 & 128 & 0.0 \\
\midrule
Command  & 4 & 128  & 0.0 \\
        & 128 & 128 & 0.0 \\
\midrule
       & 3  & 512 & 0.0 \\
Action ($a_t$) & 512  & 256 & 0.0 \\
       & 256  & 512 & 0.0 \\
       
\midrule
 Join & 512+128+128  & 512 & 0.0 \\

\midrule
 Join ($Z_t$, $Z_{t+1}$) & 512 + 512  & 512 & 0.0 \\
 
\midrule
       & 512  & 256 & 0.0\\
Action ($a_t$) & 256  & 256 & 0.5\\
       & 256  & 3 & 0.0\\
       
\midrule
Join ($Z_t$, Action) & 512 + 512  & 512 & 0.0 \\
\bottomrule
\end{tabular}} &

\resizebox{0.5\columnwidth}{!}{
\begin{tabular}{cccc}
\toprule
& Input & Output & Num. of  \\
Module & Dimension & Channels  & Dropout\\
\noalign{\smallskip}
\midrule
            & 512 & 512  & 0.0 \\
Affordances & 512 & 256 & 0.0 \\
            & 256 & M & 0.0 \\

\bottomrule
\multicolumn{4}{c}{(M: Classification - 2; Regression - 1)}\\
\end{tabular}
}\\
 & \\
(c) Forward Model & (d) Affordances Network \\
\end{tabular}
}
\end{center}
\caption{Network architecture details for the encoders $h_\theta$ (\textit{a},\textit{b} and \textit{c}) and the affordance projection  network $g_\phi$ (\textit{d}) when fine-tuned for driving.}
\label{tab:BC_configuration}
\end{table}

\section{Controller }
\label{sec:controller}
%%%%%%%%%%%%%%%%%%%%%%%

To evaluate our models in the NoCrash benchmark, we tuned a controller using our affordances. Given a set of affordances at time $t$, $\{\psi(t), hv(t), hp(t), hr(t)\}$, the controller outputs an action $a(t)$ defined by $\{S(t), T(t), B(t)\}$, {\ie}, steering, throttle, and break, respectively. In particular, for lateral control ({\ie}, $S(t)$) and for longitudinal control ({\ie}, $T(t)$ and $B(t)$) we use the following PID-based equations:
\begin{align*}
S(t) & = PID(\psi(t)) = K_p \psi(t) + K_i\int_{0}^{t} \psi(\tau) d\tau + K_d \frac{\partial\psi(t)}{\partial t}, \\
B(t) & = max(hr(t),hp(t),hv(t)), \\
T(t) & = PID(v \: \mbox{if} (B(t) > 0), \: 0 \: \mbox{otherwise}), 
\end{align*}
\noindent where the hazard functions either equal to $1$ or $0$, and $v$ is the target maximum speed, 20Km/h in these experiments. We tuned the constants $K_p, K_i$ and $K_d$ in town 1 (T1) to obtain a perfect driving (no errors, all episodes completed) for the \emph{dense} condition of the NoCrash benchmark provided we use ground truth (perfect) affordances. We did it in that way to provide a driving evaluation directly depending on the quality of the affordance predictions, not in the controller itself, since it is not the focus of this paper.

\section{Additional Results}

\mypara{Linear Probing.} Tables \ref{tbl:0.25h_frozen_T1} and \ref{tbl:0.25h_frozen_random_T1} show the linear probing evaluation results of models tested on the $\sim1$~hour Town01 testing set. This testing set has similar appearance to the training data. We observe a similar tendency to the results obtained on Town02, those shown in the main paper.

\begin{table*}[t!]
\footnotesize
\centering
\resizebox{0.95\linewidth}{!}{
\begin{tabular}{@{}lccccccccc@{}}
\toprule
&& \multicolumn{3}{c}{Binary Affordances} && \multicolumn{3}{c}{Relative Angle ($\psi_t$)} \\
 \textbf{Pre-training} && Pedestrian ($hp$) & Vehicle ($hv$) & Red T.L. ($hr$)  && Left Turn & Straight & Right Turn  \\
\midrule
No pre-training &&  $38\pm1$ &  $59\pm0$ &  $45\pm0$ &&  $10.03\pm0.06$  & $1.80\pm0.03$   & $17.57\pm0.03$\\
ImageNet &&  $35\pm1$ &  $67\pm0$ &  $54\pm1$ &&  $14.46\pm0.35$  & $2.31\pm0.09$   & $19.06\pm0.18$\\
\midrule
Contrastive (ST-DIM)  &&  $38\pm0$ &  $57\pm1$ &  $76\pm1$ &&  $6.72\pm0.03$  & $2.65\pm0.03$ & $13.12\pm0.11$\\
\midrule
Forward  &&  $\underline{56\pm0}$ &  $62\pm0$ &  $55\pm0$ &&  $6.97\pm0.12$  & $\underline{\textbf{0.17}\pm0.00}$   & $5.88\pm0.03$\\
Inverse  &&  $\underline{\textbf{57}\pm1}$ &  $\underline{82\pm0}$ &  $\textbf{89}\pm0$ &&  $\underline{3.84\pm0.06}$  & $\underline{0.21\pm0.03}$   & $\textbf{2.96}\pm0.09$\\
Behavior Cloning (BC) &&  $46\pm0$ &  $\underline{\textbf{83}\pm0}$ &  $86\pm0$ &&  $\underline{\textbf{3.76}\pm0.03}$  & $0.63\pm0.00$   & $4.35\pm0.06$\\
\bottomrule
\end{tabular}
}
\caption{Linear probing results on Town01 testing set. Left: F1 score for the binary affordances (higher is better). Results are scaled by 100 for visualization purpose. Right: MAE of the relative angle (lower is better), shown for different navigation maneuvers. MAE is shown in degrees for an easier understanding.
}
\label{tbl:0.25h_frozen_T1}
\end{table*}

\begin{table*}[!]
\footnotesize
\centering
\resizebox{0.95\linewidth}{!}{
\begin{tabular}{@{}lccccccccc@{}}
\toprule
&& \multicolumn{3}{c}{Binary Affordances} && \multicolumn{3}{c}{Relative Angle ($\psi_t$)} \\
 \textbf{Pre-training} && Pedestrian ($hp$) & Vehicle ($hv$) & Red T.L. ($hr$)  && Left Turn & Straight & Right Turn  \\

\midrule
No pre-training &&  $38\pm1$ &  $59\pm0$ &  $45\pm0$ &&  $10.03\pm0.06$  & $1.80\pm0.03$   & $17.57\pm0.03$\\
\midrule
Contrastive (ST-DIM)  &&  $36\pm1$ &  $60\pm2$ &  $67\pm1$ &&  $7.43\pm0.17$  & $2.79\pm0.18$   & $12.11\pm0.18$\\
Contrastive Random (ST-DIM)  &&  $34\pm2$ &  $\textbf{78}\pm1$ &  $52\pm2$ &&  $11.10\pm0.52$  & $2.02\pm0.09$   & $14.65\pm0.29$\\
\midrule
Forward  &&  $\textbf{52}\pm0$ &  $53\pm0$ &  $60\pm0$ &&  $3.95\pm0.00$  & $0.11\pm0.00$   & $4.37\pm0.03$\\
Forward Random  &&  $5\pm0$ &  $27\pm0$ &  $2\pm0$ &&  $19.10\pm0.03$  & $0.63\pm0.00$   & $17.97\pm0.03$\\
\midrule
Inverse  &&  $48\pm1$ &  $71\pm0$ &  $\textbf{90}\pm0$ &&  $\textbf{2.03}\pm0.07$  & $\textbf{0.21}\pm0.03$   & $\textbf{2.54}\pm0.03$\\
Inverse Random &&  $34\pm0$ &  $53\pm0$ &  $55\pm0$ &&  $12.49\pm0.45$  & $0.84\pm0.03$   & $13.22\pm0.54$\\

\bottomrule
\end{tabular}
}
\caption{Linear probing results on Town01 testing set, comparing encoders trained with random policy training data versus expert demonstration data. Note that, to provide fair comparison, the encoders here were trained with 20 hours data.}
\label{tbl:0.25h_frozen_random_T1}
\end{table*}

\begin{table}%{l}{0.50\textwidth}
\footnotesize
\centering
\resizebox{0.90\linewidth}{!}{
\begin{tabular}{@{}lccccccccc@{}}
\toprule
   && Technique  && Empty & Regular & Dense &  T.L.  \\
  \midrule

 \multirow{2}{*}{Training} && BC  driving && $72\pm5$ & $47\pm2$ & $20\pm3$ & $74\pm2$ \\
 % LBC \cite{Chen:2019} && $97\pm1$ & $93\pm1$  & $71\pm5$ & N/A && $100\pm0$ & $94\pm3$ & $51\pm3$ & N/A  \\

 &&  BC  pre-training && $\textbf{91}\pm1$ & $\textbf{91}\pm4$ & $\textbf{68}\pm5$ & $\textbf{8}\pm1$ \\
   \midrule
 \multirow{2}{*}{New Town} &&   BC driving && $71\pm2$ & $42\pm7$ & $12\pm3$ & $63\pm1$ \\
&&   BC  pre-training && $\textbf{83}\pm3$ & $\textbf{61}\pm4$ & $\textbf{25}\pm4$ & $\textbf{8}\pm1$  \\
   
\bottomrule
 \end{tabular}
}
\vspace{8pt}

\caption{We compare the behavior cloning (BC) technique  used as driving technique (BC driving) 
to it used as a pre-training technique (BC pre-training) for an
affordances based model.}
\label{tbl:compare_ete}
\end{table}

% Compared with other methods from the literature

\mypara{Driving Results.} An important observation is that using expert demonstration 
as pre-training seems to be more beneficial than training
a model end-to-end to directly perform control. 
In Table \ref{tbl:compare_ete}, we  show a behavior
cloning encoder trained with 50 hours of expert demonstrations. We compare two different uses of this encoder:  directly producing driving controls and serving as representation learning for an
affordance prediction model.
With our pre-training strategy, and the complementary affordance training, our model (BC pre-training) is able to greatly outperform the end-to-end driving results (BC driving). This difference is expressive
especially when comparing the capability to stop
on red traffic lights. Finally, note that the ``BC driving" results from
Table \ref{tbl:compare_ete} are in practice a re-training
of the CILRS \cite{Codevilla:2019} baseline to work
on version 0.9.6. 

\bibliography{example}  % .bib